\documentclass{article}

\PassOptionsToPackage{numbers, compress}{natbib}
\usepackage[preprint]{neurips_2024}

\usepackage[utf8]{inputenc} % allow utf-8 input
\usepackage[T1]{fontenc}    % use 8-bit T1 fonts
\usepackage{hyperref}       % hyperlinks
\usepackage{url}            % simple URL typesetting
\usepackage{booktabs}       % professional-quality tables
\usepackage{amsfonts}       % blackboard math symbols
\usepackage{nicefrac}       % compact symbols for 1/2, etc.
\usepackage{microtype}      % microtypography
\usepackage{xcolor}         % colors
\usepackage{graphicx}
\usepackage{amsmath}
\usepackage{adjustbox}
\usepackage{multirow}
\usepackage{wrapfig}
\usepackage{algorithm}
\usepackage{algpseudocode}
\usepackage{tabularx}
\usepackage{enumitem}
\usepackage{caption}
\usepackage{mathtools}

\allowdisplaybreaks[4]
\usepackage{multirow}
\usepackage{makecell}
\usepackage{algpseudocode}
\usepackage{longtable}
\usepackage{tabularx}
\usepackage{makecell}
\usepackage{graphicx} 
\usepackage{array,makecell}

\title{A Survey on Parameter-Efficient Fine-Tuning for Foundation Models in Federated Learning}

\author{%
  Jieming Bian  \\
  University of Florida \\
  Gainesville, FL 32611 \\
  \texttt{jieming.bian@ufl.edu} \\
\And
  Yuanzhe Peng 
  \\
  University of Florida \\
  Gainesville, FL 32611 \\
  \texttt{pengy1@ufl.edu} \\
  \And
  Lei Wang  \\
  University of Florida \\
  Gainesville, FL 32611 \\
  \texttt{leiwang1@ufl.edu} \\
  \And
  Yin Huang \\
  University of Florida \\
  Gainesville, FL 32611 \\
  \texttt{yin.huang@ufl.edu}\\
  \And
  Jie Xu \\
  University of Florida \\
  Gainesville, FL 32611 \\
  \texttt{jie.xu@ufl.edu} \\
}

\begin{document}

\maketitle

% \vspace{-20pt}

\begin{abstract}
Foundation models have revolutionized artificial intelligence by providing robust, versatile architectures pre-trained on large-scale datasets. However, adapting these massive models to specific downstream tasks requires fine-tuning, which can be prohibitively expensive in computational resources. Parameter-Efficient Fine-Tuning (PEFT) methods address this challenge by selectively updating only a small subset of parameters. Meanwhile, Federated Learning (FL) enables collaborative model training across distributed clients without sharing raw data, making it ideal for privacy-sensitive applications. This survey provides a comprehensive review of the integration of PEFT techniques within federated learning environments. We systematically categorize existing approaches into three main groups: Additive PEFT (which introduces new trainable parameters), Selective PEFT (which fine-tunes only subsets of existing parameters), and Reparameterized PEFT (which transforms model architectures to enable efficient updates). For each category, we analyze how these methods address the unique challenges of federated settings, including data heterogeneity, communication efficiency, computational constraints, and privacy concerns. We further organize the literature based on application domains, covering both natural language processing and computer vision tasks. Finally, we discuss promising research directions, including scaling to larger foundation models, theoretical analysis of federated PEFT methods, and sustainable approaches for resource-constrained environments. 
\end{abstract}

\section{Introduction}
Foundation Models (FMs)~\cite{sharegpt2023, touvron2023llama, touvron2023llama2, kim2021vilt} have demonstrated remarkable capabilities by providing robust and versatile architectures. Their ability to understand context and nuances allows them to handle diverse tasks across multiple domains, including natural language processing (NLP) and computer vision (CV). To adapt these models for specific downstream applications, fine-tuning is necessary to enhance their performance on new datasets and tasks. The conventional approach, known as full fine-tuning, involves updating all model parameters. However, this method incurs prohibitively high computational costs, particularly for large-scale models. To address this challenge, parameter-efficient fine-tuning (PEFT) methods~\cite{han2024parameter} have been proposed. These approaches selectively adjust a small subset of parameters while keeping the rest unchanged, significantly reducing computational requirements. Recent state-of-the-art PEFT methods~\cite{hu2022lora} have demonstrated that they can achieve performance comparable to full fine-tuning while substantially lowering computational costs.

\begin{figure}[h]
\centering
  \includegraphics[width=1\textwidth, trim=50 200 70 150, clip]{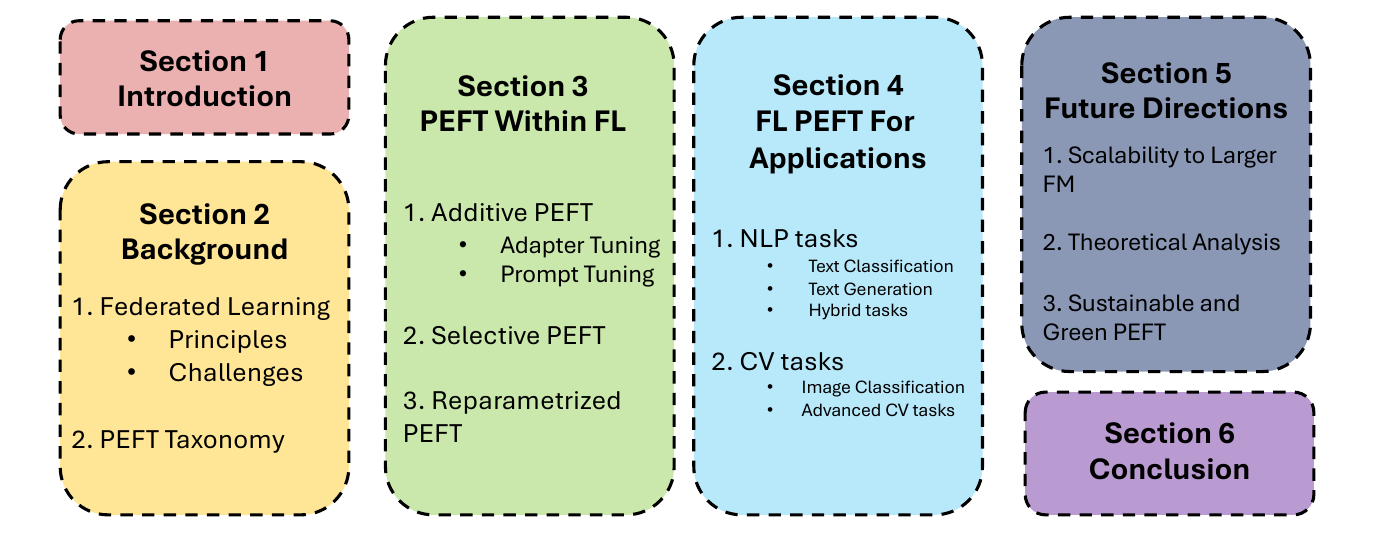}
  % \vspace{-2mm}
  \caption{A high-level roadmap of the survey structure.
  }
  \vspace{-10 pt}
  \label{fig：overall}
\end{figure}

Although PEFT methods offer significant computational savings, they still require substantial amounts of data to effectively adapt to specific downstream tasks~\cite{han2024parameter}. However, data from a single entity may be insufficient for this purpose, necessitating collaboration among multiple data owners who collectively hold the required data. This multi-owner setting raises privacy concerns, as fine-tuning across multiple parties may expose sensitive information. Privacy concerns are particularly critical in domains such as law, healthcare, and finance, where data is inherently sensitive~\cite{feretzakis2024trustworthy}. Therefore, ensuring stringent privacy safeguards is essential. Federated Learning (FL)~\cite{mcmahan2017communication} provides a viable solution by enabling collaborative model fine-tuning without requiring direct data sharing. By allowing multiple participants to contribute to model adaptation while preserving data privacy, FL effectively addresses these challenges.

Recognizing the advantages of integrating PEFT with FMs in the FL framework, numerous studies have explored this direction~\cite{wu2024fedbiot, zhao2023fedprompt}. These works leverage different PEFT methods within FL to fine-tune various FMs. The integration of FL with PEFT introduces unique challenges, such as data heterogeneity, scalability, computational constraints, and communication overhead. Researchers have proposed various solutions to address these challenges. Given the growing body of literature on PEFT in FL, a systematic survey is crucial. While several existing surveys discuss FL with FMs, none specifically focus on FL-PEFT while incorporating the latest advancements. For instance,~\cite{chen2024integration, yu2023federated} primarily examines privacy protection mechanisms in FL for FMs, while~\cite{yao2024federated, zhuang2023foundation} provide insights into the future of FMs in FL. Although~\cite{li2024synergizing} offers a detailed review of integrating FMs with FL, including discussions on PEFT, its completion date limits its coverage, as many recent FL-PEFT methods have been published after its release.

To bridge this gap, we provide a systematic and comprehensive review of parameter-efficient fine-tuning in the federated learning setting. This survey categorizes recent works based on both their PEFT methodologies and the tasks they target, such as natural language process and computer vision. 

% \begin{figure}[H]
% \centering
%   \includegraphics[width=1\textwidth]{figs/overall.png}
%   % \vspace{-2mm}
%   \caption{A high-level roadmap of the survey structure. Section 1 introduces the motivation for PEFT in FL. Section 2 outlines background concepts in FL and PEFT taxonomy. Section 3 categorizes PEFT methods within FL into additive, selective, and reparameterized types. Section 4 focuses on domain-specific FL PEFT applications in NLP and CV. Section 5 discusses future directions. Section 6 concludes the survey.
%   }
%   \label{fig：overall}
% \end{figure}

The organization of this survey is as follows. In Section~2, we introduce fundamental concepts of FL, outline its key principles, discuss common challenges, and provide an overview of PEFT. Section~3 categorizes and reviews various PEFT approaches applied in FL settings, following the taxonomy proposed in~\cite{han2024parameter}. Specifically:
Section~3.1 discusses additive PEFT methods that introduce new weight parameters or modify activations in FL. Section~3.2 examines selective PEFT methods, which fine-tune only a subset of existing parameters. Section~3.3 focuses on reparameterized PEFT methods, such as LoRA and its variants, within FL settings. Section~4 organizes the literature based on the type of pre-trained model and the targeted tasks. Section~5 outlines open research challenges and future directions for advancing the integration of PEFT in FL. Finally, in Section~6 we conclude by summarizing the findings from this survey.

\section{Background} \label{Background}
In this section, we first introduce the fundamentals of Federated Learning, outlining its key principles and discussing common challenges. We then provide an overview of various Parameter-Efficient Fine-Tuning methods.

% \begin{figure}[t]
% \centering
%   \includegraphics[width=1\textwidth]{figs/Finetuned_way.png}
%   % \vspace{-2mm}
%   \caption{An illustration of three primary model fine-tuning paradigms: (1) Local Only, where data remains solely on the data holder’s device and models are fine-tuned independently; (2) Centralized, where data from multiple sources is collected on a central server for fine-tuning a pre-trained model; and (3) Federated, where data remains decentralized and model updates are collaboratively aggregated via a central server, enabling privacy-preserving fine-tuning across distributed clients.
%   }
%   \label{fig：finetuned_way}
%   % \vspace{-18 pt}
% \end{figure}

\begin{figure}[t]
\centering
  \includegraphics[width=1\textwidth, trim=50 140 50 150, clip]{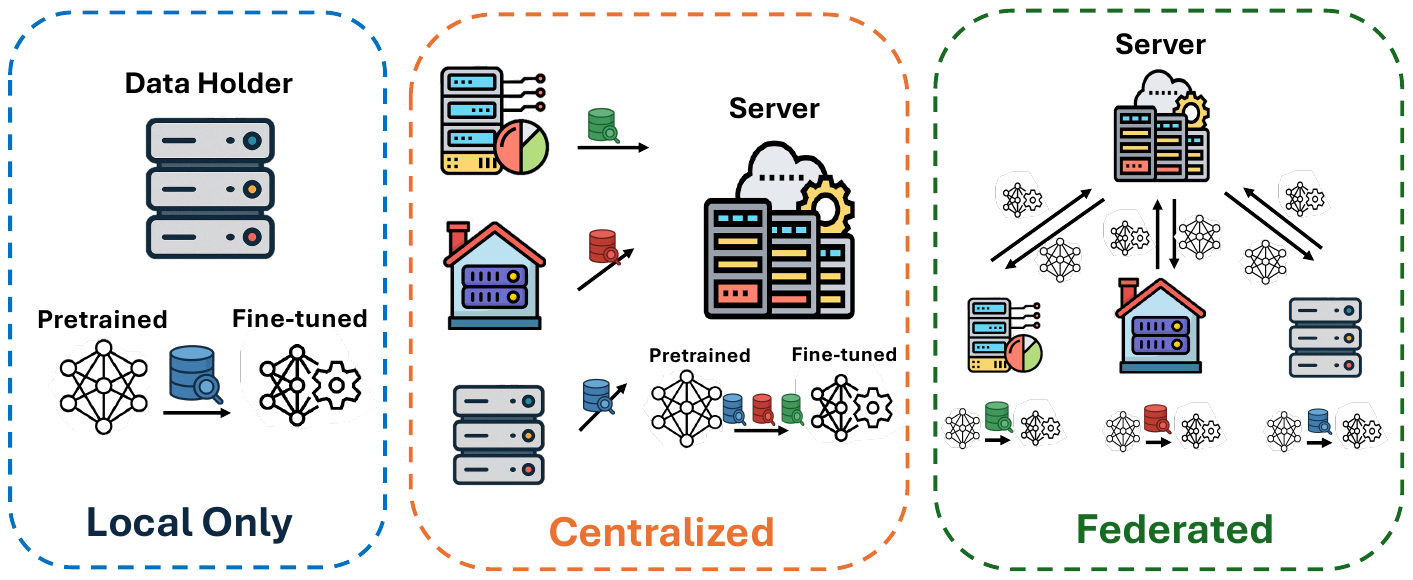}
  % \vspace{-2mm}
  \caption{An illustration of three primary model fine-tuning paradigms: (1) Local Only, where data remains solely on the data holder’s device and models are fine-tuned independently; (2) Centralized, where data from multiple sources is collected on a central server for fine-tuning a pre-trained model; and (3) Federated, where data remains decentralized and model updates are collaboratively aggregated via a central server, enabling privacy-preserving fine-tuning across distributed clients.
  }
  \label{fig：finetuned_way}
  % \vspace{-18 pt}
\end{figure}

\subsection{Federated Learning}
\subsubsection{Principles of Federated Learning}
Federated Learning is a distributed machine learning paradigm designed to preserve data privacy. Unlike traditional centralized learning, where all data is aggregated in a single location for training, FL enables models to be trained across multiple devices or organizations without exposing local data. In FL, the model moves to the data rather than the data being transmitted to the model, effectively reversing the traditional machine learning paradigm. This approach is particularly beneficial in privacy-sensitive domains such as healthcare, finance, and mobile applications.

FL operates under the following key principles: (1) \textbf{Local Model Training}: Each client trains a local model using its private data. (2) \textbf{Model Aggregation}: Instead of sharing raw data, clients send locally computed model updates to a central server. (3) \textbf{Global Model Update}: The server aggregates the local updates to refine a global model, which is then distributed back to the clients. (4) \textbf{Iterative Optimization}: This process is repeated over multiple rounds until convergence.

In a standard FL setup, multiple clients collaboratively train a global model while preserving data privacy. Each client trains on its local dataset and transmits model updates to a central server, which aggregates them to improve the global model. Consider an FL system with \( K \) clients, where training starts with an initial model \( \mathbf{W}_0 \). The global update at each round is computed as follows:

\begin{align}
\label{fl_update}
    \mathbf{\Delta W} = \sum_{k=1}^K p_k \mathbf{\Delta W}_k,
\end{align}
where \( \mathcal{D}_k \) represents client \( k \)'s local dataset, \( p_k = \frac{|\mathcal{D}_k|}{\sum_{k} |\mathcal{D}_k|} \) denotes the weight proportional to the dataset size of client \( k \), and \( \mathbf{\Delta W}_k \) is the local model update.

To initiate the next round, the global model is updated and redistributed to clients as follows:

\begin{align}
\label{fl_initial}
    &\mathbf{W}_{k,0} = \mathbf{W}_0 + \mathbf{\Delta W}; \nonumber\\
    &\mathbf{W}_{k,e+1} = \mathbf{W}_{k,e} - \eta g_{k,e}, \quad e = 0, \dots, E-1; \nonumber\\
    &\mathbf{\Delta W}_k = -\sum_{e=0}^{E-1} \eta g_{k,e},
\end{align} 
where \( \eta \) is the local learning rate, and \( g_{k,e} \) represents the stochastic gradient of client \( k \) at epoch \( e \).

\subsubsection{Challenges in Federated Learning}
\textbf{Data Heterogeneity.}
A primary challenge in FL is data heterogeneity, where data distributions across clients are non-independent and non-identically distributed (non-IID). Unlike traditional machine learning settings, where data is assumed to be IID across training samples, FL operates in decentralized environments where data collected by different clients varies due to user behaviors, preferences, and contextual factors.

Data heterogeneity in FL can be classified into two main types. \textbf{Label Distribution Skew (Label Non-IID)} occurs when clients have unequal label distributions. In centralized learning, all classes are typically well-represented in the dataset, whereas in FL, each client may only have access to a subset of labels, leading to class imbalances. For example, a hospital in one region may only collect MRI scans for a specific disease, while another hospital in a different region has data for different diseases. Mathematically, let \( P_k(y) \) denote the label distribution for client \( k \). In an IID setting, all clients share the same label distribution, i.e., \( P_1(y) = P_2(y) = \dots = P_K(y) \), whereas in real-world FL scenarios, label distributions vary, i.e., \( P_i(y) \neq P_j(y) \) for \( i \neq j \). \textbf{Feature Distribution Skew (Feature Non-IID)} arises when clients have different feature distributions, such as in speech recognition models, where users with different accents contribute to distinct audio feature distributions, or in handwritten digit recognition tasks, where variations in writing styles lead to diverse feature representations. This results in differing marginal feature distributions \( P_k(x) \) across clients, meaning \( P_i(x) \neq P_j(x) \) for \( i \neq j \).

\textbf{Computational Heterogeneity.}
In FL systems, participating clients often possess vastly different computational resources, memory capacities, and processing capabilities. These range from high-performance machines with dedicated GPUs to resource-constrained edge devices such as mobile phones, IoT devices, and embedded systems, introducing challenges in training efficiency, synchronization, and system robustness. The straggler effect occurs when clients with limited computational resources take significantly longer to complete training, causing delays in synchronous FL, where all clients must finish before aggregation. Unbalanced workloads arise because devices with different hardware capabilities require varying numbers of local iterations to achieve equivalent performance, making uniform training schedules inefficient. Additionally, memory and storage constraints limit the size of models that edge devices can train locally, further complicating FL deployment.

\textbf{Communication Efficiency.}
A major bottleneck in FL is communication efficiency, as FL requires frequent communication between clients and the central server to exchange model updates. Unlike centralized training, where raw data is transmitted to a single location, FL transmits model parameters, incurring significant communication costs. The iterative nature of FL demands multiple rounds of communication, leading to substantial bandwidth consumption. This challenge is exacerbated when training deep learning models, particularly large-scale architectures such as Large Language Models, which contain millions or even billions of parameters, making each model update highly resource-intensive.

\textbf{Privacy Concerns.}
One of the core motivations behind Federated Learning is to enhance data privacy by keeping raw data decentralized across clients. However, despite not directly sharing raw data, FL remains susceptible to various privacy threats, as adversaries can exploit shared model updates to infer sensitive information. Addressing these concerns is critical for deploying FL in privacy-sensitive domains such as healthcare, finance, and personalized services.

A key vulnerability arises from the risk of data reconstruction, where adversaries can extract original input data from shared gradients or model weights, potentially compromising user privacy. Since FL relies on exchanging gradient updates, attackers may analyze these updates to infer individual training examples, leading to gradient inversion attacks. These risks highlight the need for robust privacy-preserving techniques, such as differential privacy, secure aggregation, and homomorphic encryption, to mitigate potential threats and ensure the confidentiality of client data.

\subsection{Parameter-Efficient Fine-Tuning}
\label{sec:peft}

Large-scale pre-trained models, such as large language models (LLMs) and vision transformers (ViTs), have demonstrated remarkable performance across various domains. However, fine-tuning these models for specific downstream tasks remains essential to adapt them effectively. Full fine-tuning, which updates all model parameters, requires extensive computational resources, making it impractical for large-scale models. To address this challenge, Parameter-Efficient Fine-Tuning methods have emerged as a viable alternative, enabling task adaptation while modifying only a small subset of model parameters.

Following~\cite{han2024parameter}, we categorize PEFT techniques into three main approaches: \text{Additive PEFT}, \text{Selective PEFT}, and \text{Reparameterized PEFT}. 

\textbf{Additive PEFT} introduces additional trainable parameters while keeping the majority of the pre-trained model weights frozen. This includes methods such as adapters~\cite{houlsby2019parameter}, which insert small neural modules between transformer layers to capture domain-specific features; soft prompts~\cite{lester2021power}, which prepend learnable embeddings to the input sequence for task conditioning; and techniques like IA3~\cite{liu2022few} and SSF~\cite{lian2022scaling}, which modify intermediate activations instead of adding new trainable layers.

\textbf{Selective PEFT} fine-tunes only a subset of the model’s existing parameters while keeping the rest frozen, significantly reducing computational and storage requirements. This can be achieved through unstructured masking techniques, such as Diff Pruning and Fisher-based selection~\cite{guo2021parameter}, which dynamically identify and update only the most critical parameters, or structured masking methods like BitFit~\cite{ben-zaken2022bitfit}, which target specific parameter groups, such as bias terms or cross-attention layers, to optimize efficiency.

\textbf{Reparameterized PEFT} modifies the internal parameterization of model weights to enhance fine-tuning efficiency. A widely adopted technique in this category is Low-Rank Adaptation (LoRA)~\cite{hu2022lora}, which factorizes weight updates into low-rank matrices, significantly reducing the number of trainable parameters. Other approaches include dynamic rank adaptation methods such as DyLoRA~\cite{valipour2022dylora} and AdaLoRA~\cite{zhang2023adalora}, which adjust rank dynamically to balance efficiency and performance, as well as alternative parameterization techniques like DoRA~\cite{liu2024dora} and VeRA~\cite{kopiczko2023vera}, which explore novel weight matrix decompositions for improved task adaptation.

PEFT provides an efficient alternative to full fine-tuning, addressing the scalability challenges of large models. However, when combined with Federated Learning, additional challenges arise due to the decentralized nature and privacy constraints of FL. In the following sections, we systematically review existing research on PEFT methods within the FL setting, highlighting key advancements and challenges.

\section{PEFT within FL} \label{peft_FL}
\begin{figure}[t]
\centering
  \includegraphics[width=1\textwidth, trim=50 120 70 150, clip]{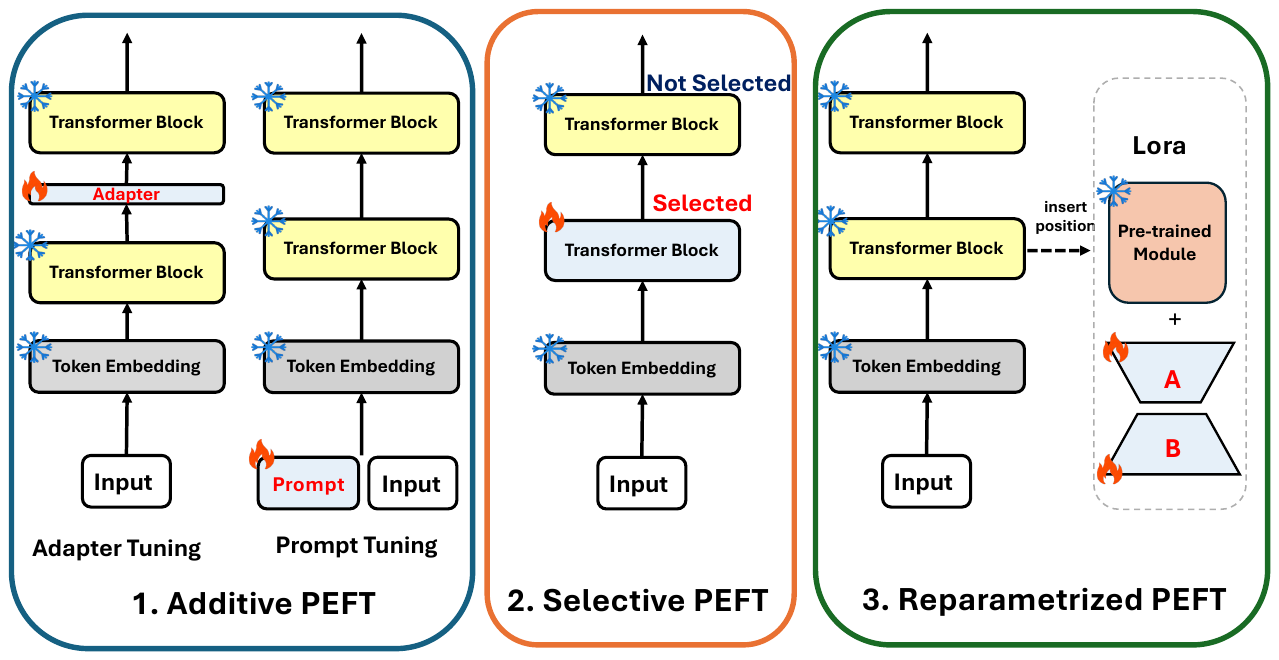}
  % \vspace{-2mm}
  \caption{A schematic comparison of three major kinds of PEFT strategies.
  }
  \label{fig：peft}
\end{figure}

Multiple types of Parameter-Efficient Fine-Tuning methods have been developed in recent years. In this survey, we follow the categorization defined by \cite{han2024parameter} and focus on PEFT methods specifically developed for the federated learning setting. Throughout this section, we examine three primary categories of PEFT in FL: Additive PEFT, Selective PEFT, and Reparameterized PEFT. A schematic comparison of the three major PEFT strategies (Additive, Selective, and Reparameterized) is presented in Fig.\ref{fig：peft}.

\subsection{Additive PEFT}
To reduce the computational costs associated with full parameter fine-tuning, a widely employed approach is to keep foundation models fixed while introducing a minimal number of trainable parameters strategically positioned within the model. During fine-tuning for specific downstream tasks, only the weights of these additional modules or parameters are updated, resulting in substantial reductions in storage, memory, and computational resource requirements. Key branches within additive methods include adapter tuning and prompt tuning.

\subsubsection{Adapter Tuning}
Adapter tuning in Federated Learning introduces small trainable modules within Transformer blocks, enabling efficient task adaptation while keeping most model parameters frozen. This approach significantly enhances both computational and communication efficiency, making it particularly suitable for FL, where resource constraints and data privacy are critical concerns. Various FL-specific adapter tuning strategies have been developed, each addressing key challenges such as data heterogeneity and communication overhead.

A primary challenge in FL is data heterogeneity, where variations in client data distributions can lead to suboptimal model performance. To address this, several methods leverage client-specific adapter designs. FedDAT~\cite{chen2024feddat} introduces a Dual-Adapter Teacher framework that enhances multi-modal FL by employing Mutual Knowledge Distillation to regularize client updates and facilitate knowledge transfer, particularly in vision-language tasks. C2A~\cite{kim2023client} introduces a hypernetwork-based FL framework that generates personalized adapters based on each client’s data distribution. By conditioning on both label distribution and contextual embeddings, C2A dynamically adjusts adapter parameters while factorizing hypernetworks to reduce the number of trainable parameters. Adapter-based personalization has also been explored in specialized applications such as federated medical analysis, where~\cite{zhao2023multi} fine-tunes pre-trained medical models using task adapters, enabling personalized multi-task FL while preserving data privacy.

Beyond personalization, adapter tuning also facilitates cross-client knowledge sharing across clients with different tasks or domains. Pilot~\cite{xiong2025pilot} proposes a two-stage "adapter-on-adapter" framework that first extracts task-specific and client-specific visual features and then enables cross-task knowledge sharing through a Cross-task Mixture-of-Adapters (CT-MoA) module. Additionally, it employs an adaptive parameter aggregation strategy to minimize conflicts across tasks. ADAFEDSELECKD~\cite{feng2024adapter} further enhances FL adaptability by introducing a dual-adapter architecture consisting of global and local adapters, where local adapters undergo an entropy-based selective knowledge distillation process to filter uncertain global knowledge while optimizing for domain-specific performance. MPFT~\cite{zhang2024enhancing} takes a different approach by introducing a prototype-based federated fine-tuning framework that facilitates knowledge transfer without raw data exchange. By leveraging multi-domain prototypes—pretrained representations enriched with domain-specific information—the framework enables more effective adaptation across clients. 

Communication efficiency is another critical aspect of adapter tuning, as it inherently reduces the number of trainable parameters exchanged between clients. Several studies have further optimized this advantage to minimize communication costs. Fed-MNMT~\cite{liu2023communication} addresses this challenge by employing clustering strategies based on language families, gradient similarity, or random assignment, ensuring that clients with similar linguistic characteristics share adapter updates more effectively. This method significantly reduces communication overhead by over 98\%.

To tackle computational heterogeneity among clients, FedAdapter~\cite{cai2022fedadapter} addresses computational heterogeneity in FL-based NLP through a novel progressive training approach, wherein it strategically begins with small adapters at upper layers to efficiently learn shallow knowledge with minimal resource consumption. Subsequently, as training progresses, it intelligently expands to deeper layers only when necessary, thereby dynamically adapting to diverse computational capabilities across devices. Furthermore, FedAdapter employs innovative sideline trials where participant devices are allocated to different configuration groups, enabling continuous profiling of potential adapter configurations and precise determination of optimal upgrade timing based on actual device performance characteristics.

\subsubsection{Prompt Tuning}

Prompt tuning is an alternative parameter-efficient fine-tuning strategy where a small set of continuous, trainable vectors (soft prompts) are prepended to input embeddings, while the underlying model parameters remain frozen. Unlike adapter tuning, prompt tuning requires no architectural modifications to the model, making it lightweight and easily deployable. In the context of Federated Learning, prompt tuning is especially appealing due to its minimal communication overhead—only the compact prompt parameters are shared between clients and the server—alongside strong privacy preservation, as prompts do not directly reveal raw data.

Communication efficiency is a central advantage of prompt tuning in FL. PromptFL~\cite{pan2024federated} introduces one of the earliest frameworks in this space by training shared soft prompts on foundation models like CLIP, substantially reducing global aggregation overhead and improving performance with limited local data. FedPrompt~\cite{zhao2023fedprompt} enhances communication efficiency through split aggregation of soft prompts, making the method more robust to backdoor attacks. GFPT~\cite{wang2024personalized} proposes a discrete local search approach using natural language tokens as prompts, replacing backpropagation through the entire model. It further compresses uploads and downloads using token indices and compressed embeddings, reducing memory and communication costs while maintaining competitive performance on text classification tasks.

Client data heterogeneity—in terms of data distribution, domain, or task—is a persistent challenge in FL. To address this, a number of works adopt personalized or domain-adaptive prompt tuning strategies. pFedPrompt~\cite{guo2023pfedprompt} personalizes soft prompts for each client in multimodal vision-language models like CLIP, allowing adaptation to unique visual features. Similarly, FedAPT~\cite{su2024federated} introduces adaptive prompt tuning by dynamically assigning domain-specific prompts based on test-time input characteristics, using meta-prompts and an adaptive network module to facilitate collaborative learning across domains.

Extending this direction, pFedMoAP~\cite{luo2024mixture} leverages a mixture-of-adaptive-prompts framework, fusing local and non-local prompt experts using a local attention-based gating network. DiPrompT~\cite{bai2024diprompt} disentangles global knowledge (via global prompts) from domain-specific features (via domain prompts), enabling flexible adaptation during inference. FedOTP~\cite{li2024global} employs optimal transport theory to balance local and global personalization via adaptive transport plans between prompts. pFedPG~\cite{yang2023efficient} proposes server-side personalized prompt generators that encode client-specific optimization directions to better tailor local prompts.

To balance generalization and personalization, several studies introduce hybrid prompt strategies. FedHPL~\cite{ma2024fedhpl} combines local visual prompt tuning with global logit distillation to enable effective cross-client knowledge transfer, even with heterogeneous model architectures. PromptFolio~\cite{pan2024federated} proposes a `prompt portfolio' mechanism and provides theoretical and empirical analysis of how local and global prompts can be optimally combined to achieve the best trade-off between generalization and personalization. SGPT~\cite{deng2024unlocking} introduces both shared and group-specific prompts, using a selection module to adaptively choose prompts based on local data distributions, optimized through block coordinate descent to prioritize general knowledge before specialization.

Further, Application-specific adaptations of prompt tuning have also emerged. In recommendation systems, PFCR~\cite{guo2024prompt} proposes a two-stage FL framework: it first learns domain-agnostic content representations using Federated Content Representation (FCR) in a privacy-preserving way, followed by domain-specific prompting (either full or lightweight) for fine-tuning without requiring overlapping users across domains. In federated continual learning, Powder~\cite{piao2024federated} uses dual prompt-based mechanisms for both spatial (cross-client) and temporal (across-task) knowledge transfer, addressing catastrophic forgetting while maintaining low communication overhead.

Low computational resource adaptation is another area where prompt tuning excels. To support regions with limited training data and constrained computational resources,~\cite{zhao2024breaking} introduces federated prompt averaging to enable effective cross-lingual knowledge transfer. By leveraging global prompt information across clients speaking different languages, the method improves accuracy in low-resource settings without compromising privacy or requiring intensive computation.

A summary of federated learning methods that employ additive PEFT approaches is provided in Table \ref{tab:fl_additive}, offering a comparative overview of their key characteristics and contributions to addressing various FL challenges.

\begin{longtable}{>{\centering\arraybackslash}m{0.08\linewidth} >{\centering\arraybackslash}m{0.17\linewidth} >{\centering\arraybackslash}m{0.27\linewidth} >{\centering\arraybackslash}m{0.38\linewidth}}
    \caption{Federated Learning with Additive PEFT.}
    \vspace{10 pt}
    \label{tab:fl_additive} \\
    \hline
    \textbf{Specific} & \textbf{Method} & \textbf{FL Challenge} & \textbf{Method Highlight} \\
    \hline
    \endfirsthead
    
    % 继续表头（翻页时重复）
    \hline
    \textbf{Specific} & \textbf{Method} & \textbf{FL Challenge} & \textbf{Method Highlight} \\
    \hline
    \endhead
    
    % 表格内容 - Adapter Methods
    Adapter & FedDAT \cite{chen2024feddat} & Communication Efficiency, Data Heterogeneity & Dual-Adapter Teacher and Mutual Knowledge Distillation \\
    \cline{2-4}
    & FedAdapter \cite{cai2022fedadapter} & Communication Efficiency, Computational Heterogeneity & Progressive adapter configuration with auto-scaling depth/width \\
    \cline{2-4}
    & Fed-MNMT \cite{liu2023communication} & Communication Efficiency & Clustering strategies to group clients \\
    \cline{2-4}
    & C2A \cite{kim2023client} & Data Heterogeneity & Hypernetwork generates client-customized adapters \\
    \cline{2-4}
    & MPFT \cite{zhang2024enhancing} & Data Heterogeneity & Uses multi-domain prototypes to fine-tune without averaging \\
    \cline{2-4}
    & Pilot \cite{xiong2025pilot} & Data Heterogeneity & Two-stage adapter architecture with Cross-task Mixture-of-Adapters for multi-modal \\
    \cline{2-4}
    & ADAFED \cite{feng2024adapter} & Data Heterogeneity & Dual-adapter architecture with entropy-based selective knowledge distillation \\
    \cline{2-4}
    & MT-FedAdapter~\cite{zhao2023multi} & Data Heterogeneity & Task-specific adapters with defined relationships \\
    \hline
    
    % Prompt Methods
    Prompt & GF-PT \cite{wang2024personalized} & Communication Efficiency & Gradient-free discrete local search with token-based compression \\
    \cline{2-4}
    & FedPrompt \cite{zhao2023fedprompt} & Communication Efficiency, Privacy Preservation & Basic method which only tunes and aggregates soft prompts \\
    \cline{2-4}
    & PromptFL \cite{guo2023promptfl} & Communication Efficiency & Ships frozen CLIP to clients who collaboratively train soft prompts \\
    \cline{2-4}
    & FedHPL \cite{ma2024fedhpl} & Computational Heterogeneity, Data Heterogeneity & Weighted knowledge aggregation with logit distillation \\
    \cline{2-4}
    & pFedPG \cite{yang2023efficient} & Computational Heterogeneity, Data Heterogeneity & Client-specific prompt generator produces personalized prompts from optimization directions \\
    \cline{2-4}
    & Powder \cite{piao2024federated} & Continual Learning, Data Heterogeneity & Two-step aggregation with task correlation estimation and dual distillation loss \\
    \cline{2-4}
    & pFedMoAP \cite{luo2024mixture} & Data Heterogeneity & Attention-based gating network with local and non-local prompt experts \\
    \cline{2-4}
    & FedAPT \cite{su2024federated} & Data Heterogeneity & Adaptive network generates domain-aware personalized prompts \\
    \cline{2-4}
    & pFedPrompt \cite{guo2023pfedprompt} & Data Heterogeneity & Dual-modality approach with global consensus and local feature attention \\
    \cline{2-4}
    & PFCR \cite{guo2024prompt} & Data Heterogeneity, Privacy Preservation & Two-stage: FCR learning and adaptive prompting \\
    \cline{2-4}
    & DiPrompT \cite{bai2024diprompt} & Data Heterogeneity & Disentangled global and domain-specific prompts with dynamic query \\
    \cline{2-4}
    & SGPT \cite{deng2024unlocking} & Data Heterogeneity & Shared and group prompts to bridge generalized and personalized FL \\
    \cline{2-4}
    & FedOTP \cite{li2024global} & Data Heterogeneity & Dual-prompt learning with unbalanced Optimal Transport for global consensus and local personalization \\
    \cline{2-4}
    & PromptFolio \cite{pan2024federated} & Data Heterogeneity & Global-local prompt portfolio with optimal mixing coefficient derived from feature learning theory \\
    \cline{2-4}
    & MFPT \cite{zhao2024breaking} & Data Heterogeneity, Privacy Preservation & Lightweight prompt encoders for cross-lingual transfer with low-resource languages \\
    \hline
\end{longtable}%

\subsection{Selective PEFT}

Selective Parameter-Efficient Fine-Tuning focuses on fine-tuning only a subset of a model’s existing parameters rather than adding new components, as in additive PEFT. This approach reduces computational and memory costs while still enabling effective adaptation to downstream tasks. In contrast to additive methods that increase model complexity by introducing additional parameters, selective PEFT aims to achieve efficiency by modifying only critical components of the model.

In the broader context of traditional federated learning (i.e., not focused on foundation models), selective training has been extensively studied. For example, CELL~\cite{seo2021communication} proposes a communication-efficient personalized FL method by broadcasting full models rather than unicasting subnetworks. It dynamically adjusts validation thresholds for stragglers and alternates between lottery-based pruning and FL modes depending on user performance. HeteroFL~\cite{diao2020heterofl} enables efficient training of heterogeneous local models with varying computation capacities by distributing subnetworks according to client resources. Despite the diversity in local model sizes, it still produces a unified global model through a specialized aggregation strategy that ensures fair knowledge transfer from clients with smaller models.

When moving into the domain of fine-tuning foundation models under FL, selective PEFT strategies have been adapted and extended to meet the challenges unique to this setting. Several approaches build directly on selective PEFT methods originally proposed for centralized learning. For instance, FedPEFT-Bias~\cite{sun2024exploring} selectively fine-tunes only the bias terms of pre-trained vision models, keeping all other weights frozen. This minimalistic yet effective strategy allows efficient personalization in FL with limited resource consumption. Similarly, FedBF~\cite{zhang2023fedpetuning} integrates the BitFit method~\cite{zaken2021bitfit}—which fine-tunes only bias terms—into a federated setting, enabling privacy-preserving and efficient training for NLP applications.

Some works in FL use Selective PEFT to address data heterogeneity challenge in FL. FedSelect~\cite{tamirisa2023fedselect} introduces a parameter-selective personalization strategy based on the Gradient-based Lottery Ticket Hypothesis. It dynamically identifies which parameters to fine-tune locally and which to share globally, simultaneously discovering the optimal split for personalization and global knowledge sharing. This removes the need for manual selection of tunable layers and supports flexible, client-specific architectures. FedPCL~\cite{tan2022federated} takes a different approach by leveraging prototype-wise contrastive learning. It uses class prototypes as knowledge carriers between clients and the server, extracting class-relevant representations through contrastive learning with both local and global prototypes to facilitate effective knowledge transfer.

To address computational heterogeneity across clients, several methods scale models adaptively to match client resource constraints. DepthFL~\cite{kim2023depthfl} proposes a depth-wise model scaling approach where local models are constructed by pruning the deeper layers of a global model. It uses mutual self-distillation between classifiers of varying depths, allowing shallow classifiers (trained on more data from resource-constrained clients) to guide deeper ones. However, because deeper layers are only trained on a small subset of high-resource clients, this may lead to biased feature representations. To mitigate this issue, FedRA~\cite{su2024fedra} introduces a random allocation strategy, where clients are randomly assigned different subsets of the model’s layers in each communication round. This ensures that all layers receive updates from a diverse set of clients, helping address feature imbalance. Additionally, FedRA reorganizes subsets of the global model to support resource-constrained clients and applies adapter-based fine-tuning, followed by aggregating updated adapter parameters according to a layer allocation matrix. This method enables full-model training across clients, even when no individual client has the capacity to train the entire model. A summary of federated learning methods that employ selective PEFT approaches is provided in Table \ref{tab: fl_selective}.

\begin{longtable}{>{\centering\arraybackslash}m{0.17\linewidth} >{\centering\arraybackslash}m{0.27\linewidth} >{\centering\arraybackslash}m{0.38\linewidth}}
    \caption{Federated Learning with Selective PEFT.}
    \vspace{10 pt}
    \label{tab: fl_selective} \\
    \hline
    \textbf{Method} & \textbf{FL Challenge} & \textbf{Method Highlight} \\
    \hline
    \endfirsthead

    \hline
    \textbf{Method} & \textbf{FL Challenge} & \textbf{Method Highlight} \\
    \hline
    \endhead

    % Empty rows for future fill-in
    DepthFL \cite{kim2023depthfl} & Computational Heterogeneity & Depth-scaled local models with mutual self-distillation between multiple classifiers \\
    \cline{1-3}
    FedRA \cite{su2024fedra} & Computational Heterogeneity, Data Heterogeneity & Random Allocation strategy ensures every model layer learns from all clients \\
    \cline{1-3}
    FedSelect \cite{tamirisa2023fedselect} & Computational Heterogeneity, Data Heterogeneity & Parameter-selective personalization via Gradient-based Lottery Ticket Hypothesis \\
    \cline{1-3}
    FedPCL \cite{tan2022federated} & Communication Efficiency, Data Heterogeneity & Prototype-wise contrastive learning with global and local class prototypes for knowledge sharing\\
    \cline{1-3}
    FedPEFT-Bias \cite{sun2024exploring} & Communication Efficiency & Parameter-efficient bias-only tuning with frozen weights to enable large foundation models \\
    \cline{1-3}
    FedBF \cite{zhang2023fedpetuning} & Communication Efficiency, Privacy Preservation & Combine FedAvg with BitFit to reduce resource costs \\
    \cline{1-3}
    \hline
\end{longtable}

\subsection{Reparameterized PEFT}
Reparameterization involves transforming a model's architecture while preserving its functional equivalence through parameter transformation. In the context of PEFT, this typically entails constructing a low-rank parameterization to achieve parameter efficiency during training \cite{han2024parameter}.

The most widely recognized reparameterization technique is Low-Rank Adaptation (LoRA) \cite{hu2022lora}. LoRA substantially reduces the number of trainable parameters in large-scale models by incorporating low-rank matrices into the model. Consider a pre-trained model with parameters $\mathbf{W}_0 \in \mathbb{R}^{d \times l}$, where $\mathbf{W}_0$ represents the fixed parameters of the model, and $\mathbf{\Delta W} \in \mathbb{R}^{d \times l}$ denotes the trainable update matrix applied during fine-tuning. Rather than updating all elements in $\mathbf{\Delta W}$, LoRA decomposes $\mathbf{\Delta W}$ into two low-rank matrices $\mathbf{A} \in \mathbb{R}^{r \times l}$ and $\mathbf{B} \in \mathbb{R}^{d \times r}$, where $r \ll \min(d, l)$. Thus, the model update is expressed as $\mathbf{\Delta W} = \mathbf{B} \mathbf{A}$, allowing the fine-tuning process to focus on the much smaller low-rank matrices $\mathbf{A}$ and $\mathbf{B}$ instead of the full matrix $\mathbf{\Delta W}$. Consequently, the total number of parameters that need to be trained is reduced from $d \times l$ to $r \times (d + l)$, where $r$ is significantly smaller than both $d$ and $l$. The updated model parameters after fine-tuning are given by:

\begin{align}\label{lora} \mathbf{W} = \mathbf{W}_0 + \mathbf{\Delta W} = \mathbf{W}_0 + \mathbf{B} \mathbf{A}. \end{align}

In practice, $\mathbf{A}$ is typically initialized with random Gaussian values, while $\mathbf{B}$ is initialized to zero to ensure stability at the beginning of the fine-tuning process. This low-rank adaptation enables LoRA to achieve performance comparable to full fine-tuning while significantly reducing computational and memory overhead.

Given that LoRA can achieve performance comparable to full parameter fine-tuning, federated fine-tuning with LoRA has recently emerged as an active research area. A fundamental method that directly combines LoRA with FL is FedIT \cite{zhang2024towards}. In FedIT, each of the $K$ clients begins with a fixed pre-trained foundation model $\mathbf{W}_0$ and trains the local LoRA modules represented as low-rank matrices \( \mathbf{A}_k \) and \( \mathbf{B}_k \) on its private dataset $\mathcal{D}_k$. The server then aggregates these local matrices uploaded by clients into global LoRA modules, $\mathbf{\bar{A}}$ and $\mathbf{\bar{B}}$, through a weighted average based on data size:
\begin{align}
\label{global_modules}
\mathbf{\bar{A}} = \sum_{k=1}^K p_k \mathbf{A}_k, \quad \mathbf{\bar{B}} = \sum_{k=1}^K p_k \mathbf{B}_k,
\end{align}
where \( p_k = \frac{|\mathcal{D}_k|}{\sum_{k=1}^K |\mathcal{D}_k|} \) reflects each client's data proportion. Using these averaged matrices, the server distributes them back to the clients for subsequent training rounds. In FedIT, the actual global update received by each client is:
\begin{align}
\label{eq:approx_update}
\mathbf{\Delta W}' = \mathbf{\bar{B}} \mathbf{\bar{A}} = \left( \sum_{k=1}^K p_k \mathbf{B}_k \right) \left( \sum_{k=1}^K p_k \mathbf{A}_k \right).
\end{align}
However, this aggregated update deviates from the ideal global model update in the typical FL setting, which should be the weighted sum of all local model updates:
\begin{align}
\label{eq:ideal_update}
\mathbf{\Delta W} = \sum_{k=1}^K p_k \mathbf{\Delta W}_k = \sum_{k=1}^K p_k \mathbf{B}_k \mathbf{A}_k \neq \mathbf{\Delta W}'.
\end{align}

The reparameterized nature of LoRA fine-tuning introduces a unique challenge when applied to federated learning settings, which we refer to as \textbf{Server-Side Aggregation Bias}. To address this bias, several federated learning methods with LoRA have been proposed. FFA-LoRA \cite{sun2024improving} suggests fixing randomly initialized non-zero matrices $\mathbf{{A}}$ and only fine-tuning zero-initialized matrices $\mathbf{{B}}$. However, fine-tuning only the matrices $\mathbf{{B}}$ can limit the final model performance. To improve this, LoRA-A² \cite{koo2024towards} proposes alternating freezing between LoRA modules A and B across communication rounds. Similarly, RoLoRA \cite{chen2025robust} proposes alternating optimization to fine-tune LoRA adapters in federated learning. FLoRA \cite{wang2024flora} introduces a stacking-based aggregation mechanism for federated fine-tuning with LoRA that enables noise-free aggregation of local LoRA modules and naturally supports heterogeneous LoRA ranks. Although FLoRA eliminates the server-side aggregation bias, this method incurs high communication costs proportional to the number of clients and raises privacy concerns, as it requires distributing all clients' LoRA modules to each client rather than only the averaged modules. To address this issue, FedEx-LoRA \cite{singhal2024exact} proposes adding a residual error term to the pretrained frozen weight matrix that captures the discrepancy between the average of products and the product of averages in federated LoRA aggregation. This method requires computing SVD each round, which increases server-side computational cost. To mitigate this, LoRA-FAIR \cite{bian2024lora} proposes a novel federated fine-tuning approach that simultaneously addresses server-side aggregation bias and client-side initialization lag in LoRA-based federated learning. It introduces a residual correction term to the LoRA module B at the server, minimizing the discrepancy between the approximated and ideal global updates while preserving continuity between training rounds through regularization.

Data heterogeneity, as one of the most significant challenges in FL settings, has prompted several works with LoRA fine-tuning to address this problem. pFedLoRA \cite{yi2023pfedlora}, which does not focus on foundation model fine-tuning, first introduced the idea of using LoRA for personalized FL. It proposes a mutual learning framework where each client's heterogeneous model is augmented with a small homogeneous low-rank adapter, enabling an iterative training process where the adapter and the local model take turns being frozen and updated to efficiently transfer global and personalized knowledge, with only the adapter parameters transmitted for server aggregation. FedSA-LoRA \cite{guo2024selective} proposes a selective aggregation approach where two low-rank trainable matrices (A and B) model weight updates, but only A matrices that capture general knowledge are shared with the server for aggregation, while B matrices that model client-specific knowledge remain local, enabling personalization while maintaining communication efficiency. PF2LoRA \cite{hao2025personalized} proposes a two-level LoRA framework using bilevel optimization to learn a common adapter shared by all clients while simultaneously enabling client-specific personalization through lightweight adapters. FRLoRA \cite{yanfederated} proposes a residual low-rank adaptation approach that directly sums the weight of global model parameters with a residual low-rank matrix product during the global update step and reinitializes the local low-rank matrices with the principal singular values and vectors of pre-trained weights in each round to calibrate inconsistent convergence. FedP2EFT \cite{lee2025fedp} proposes a federated learning-to-personalize approach that collaboratively trains a personalization strategy generator (PSG) across clients to dynamically determine client-specific optimal LoRA rank structures based on local meta-data and uses Bayesian sparse rank selection to efficiently allocate computation and communication resources while maintaining high performance on multilingual data.

The above personalized LoRA fine-tuning methods primarily focus on milder heterogeneity settings where clients share the same underlying task but differ in label distribution. However, in real-world LLM deployment, clients often have different tasks, presenting unique challenges that prior methods are not designed to handle. FedDPA \cite{long2024dual} proposes a dual-personalizing adapter architecture that simultaneously tackles test-time distribution shifts and client-specific personalization, considering that each client has heterogeneous tasks. To further address how conventional aggregation-based FL suffers from harmful cross-client interference in heterogeneous settings, FedALT \cite{bian2025fedalt} introduces a decoupled training scheme where clients maintain Individual LoRA and RoTW LoRA, with the latter kept frozen during training. This departs from the FedAvg paradigm underpinning existing approaches, replacing global model aggregation with frozen RoTW LoRA to share knowledge while preserving local adaptability. FedALT incorporates a novel adaptive mixer, inspired by MoE, tailored to federated personalization. It learns input-specific weights between Individual and RoTW components, enabling precise balancing of personalization and global knowledge. FedOA \cite{longfederated} introduces an additional personalized model with feature distance-based regularization from the global model, specializing while leveraging global invariant features to enhance OOD generalization.

In addition to data heterogeneity, several works address computational heterogeneity in federated LoRA fine-tuning. HETLORA \cite{cho2024heterogeneous} proposes a heterogeneous-rank LoRA approach for federated fine-tuning that allows different rank LoRA modules across devices. By applying rank self-pruning locally and sparsity-weighted aggregation at the server, HETLORA combines the advantages of high and low-rank LoRAs, achieving improved convergence speed and final performance with enhanced computation efficiency compared to homogeneous LoRA and full fine-tuning methods. FlexLoRA \cite{bai2024federated} proposes a simple yet effective aggregation scheme that enables mixing diverse LoRA weights across heterogeneous clients. By synthesizing a full-size LoRA weight from individual client contributions and employing SVD for weight redistribution, FlexLoRA fully leverages heterogeneous client resources to enhance the generalization ability of the global model while maximizing computational efficiency compared to traditional FL methods. Similarly, SLoRA \cite{babakniya2023slora} proposes a two-stage parameter-efficient fine-tuning approach that overcomes the limitations of LoRA in federated settings with heterogeneous data. It first employs a sparse fine-tuning phase using a server-generated random mask to reduce communication costs, followed by SVD-based decomposition of accumulated updates to initialize LoRA blocks. LEGEND \cite{liu2024adaptive} proposes an adaptive LoRA-based federated fine-tuning framework that determines device-specific LoRA depth and rank distribution based on computing and communication capabilities. The method uses a greedy-based algorithm to calculate the gap between maximum and minimum LoRA depth, then assigns appropriate depths to devices based on their relative capabilities.

Beyond addressing heterogeneity issues, some papers tackle other practical challenges in federated fine-tuning with LoRA. For example, FedBiOT \cite{wu2024fedbiot} considers the setting where the most recent and powerful versions of LLMs are usually closed-source. In this approach, the server generates a compressed model and divides it into an emulator (that simulates the original raw model) and an adapter (that learns domain-specific patterns). Clients only fine-tune the lightweight adapter using LoRA, significantly reducing resource consumption, while the server distills the emulator from the original LLM and aggregates the updated adapters. A bi-level optimization framework minimizes the negative effect of data distribution shift between server and clients. A summary of federated learning methods that employ reparameterized PEFT approaches is provided in Table \ref{tab:fl_lora}.

\begin{longtable}{>{\centering\arraybackslash}m{0.17\linewidth} >{\centering\arraybackslash}m{0.32\linewidth} >{\centering\arraybackslash}m{0.38\linewidth}}
    \caption{Federated Learning with Reparameterized PEFT.}
    \vspace{10 pt}
    \label{tab:fl_lora} \\
    \hline
    \textbf{Method} & \textbf{FL Challenge} & \textbf{Method Highlight} \\
    \hline
    \endfirsthead

    \hline
    \textbf{Method} & \textbf{FL Challenge} & \textbf{Method Highlight} \\
    \hline
    \endhead

    % Empty rows for future fill-in
    FFA-LoRA \cite{sun2024improving} &  Privacy Preservation, Server-Side Aggregation Bias & Freezes initialized A matrices and only fine-tunes B matrices  \\
    \cline{1-3}
    LoRA-A$^2$ \cite{koo2024towards} & Computational Heterogeneity, Server-Side Aggregation Bias  & Alternating freeze between LoRA matrices and adaptive rank selection \\
    \hline
    RoLoRA \cite{chen2025robust} & Server-Side Aggregation Bias & Alternating optimization of LoRA adapters \\
    \hline
    FLoRA \cite{wang2024flora} & Computational Heterogeneity, Server-Side Aggregation Bias  & Stacking-based aggregation of LoRA adapters to enable heterogeneous ranks and noise-free model updates \\
    \hline
    FedEx-LoRA \cite{singhal2024exact} & Server-Side Aggregation Bias & Residual error term for exact aggregation with minimal overhead \\
    \hline
    LoRA-FAIR \cite{bian2024lora} & Server-Side Aggregation Bias & Server-side residual correction with regularization for exact aggregation \\
    \hline
    FedSA-LoRA \cite{guo2024selective} & Data Heterogeneity & Selectively aggregates only A matrices that capture general knowledge while keeping B matrices that model client-specific knowledge\\
    \hline
     PF2LoRA \cite{hao2025personalized} & Data Heterogeneity & Two-level LoRA with automatic rank learning  \\
     \hline
     FedDPA \cite{long2024dual} & Data Heterogeneity & Dual-personalizing adapter with instance-wise dynamic weighting \\
     \hline
     FedALT \cite{bian2025fedalt} & Data Heterogeneity & Individual and Rest-of-the-World components combined via adaptive MoE mixer \\
     \hline
     FedOA \cite{longfederated} & Data Heterogeneity, Out-of-Distribution Generalization & Feature distance-based regularization for aligning personalized and global models \\
     \hline
    FRLoRA \cite{yanfederated} & Data Heterogeneity & Residual low-rank matrix product and principal singular space reinitialization \\
    \hline
    FedP2EFT \cite{lee2025fedp} & Data Heterogeneity & Federated learning-to-personalize approach with a PS generator for client-tailored LoRA rank selection \\
    \hline
    HETLORA \cite{cho2024heterogeneous} & Computational Heterogeneity, Data Heterogeneity & Heterogeneous rank LoRA with rank self-pruning and sparsity-weighted aggregation \\
    \hline
    FlexLoRA \cite{bai2024federated} & Computational Heterogeneity, Data Heterogeneity & Aggregation of heterogeneous-rank LoRA modules via SVD-based weight redistribution \\
    \hline
    SLoRA \cite{babakniya2023slora} & Computational Heterogeneity, Communication Efficiency & Two-stage approach with sparse fine-tuning and LoRA initialization via SVD decomposition \\
    \hline
    LEGEND \cite{liu2024adaptive} & Computational Heterogeneity & Adaptive LoRA depth and rank distribution for heterogeneous devices with gradually increasing rank for deeper layers \\
    \hline
    FedBiOT \cite{wu2024fedbiot} & Closed Source FM & Bi-level optimization for LLM fine-tuning with compressed model and LoRA-based adapters \\
    \hline
\end{longtable}

\section{FL PEFT for Applications } \label{applications}
This section presents a comprehensive analysis of Parameter-Efficient Fine-Tuning applications in federated learning environments, organized by primary task domains. Rather than categorizing by specific PEFT methodologies, we focus on how these techniques apply to different application scenarios, primarily in Natural Language Processing and Computer Vision. For each domain, we examine suitable datasets for federated learning scenarios, recommended foundation models that can serve as baselines, and relevant PEFT methods that have demonstrated promising results in federated settings. The summary can be found in Tables. \ref{tab: nlp} and \ref{tab:fl_peft_cv}.

\subsection{Natural Language Processing Tasks}
\subsubsection{Text Classification}
Text classification represents one of the most extensively studied tasks in federated learning with PEFT methods. This domain encompasses sentiment analysis, topic classification, natural language inference, and other specialized classification tasks, all of which present unique challenges in federated environments due to varying data distributions across clients.

From a dataset perspective, several collections have emerged as standard benchmarks for evaluating federated text classification approaches. General text classification datasets such as 20Newsgroup \cite{lang1995newsweeder} and AGNEWS \cite{zhang2015character} provide diverse topical categories that can be naturally partitioned to simulate non-IID data distributions across clients. For sentiment analysis, the SST-2 \cite{socher2013recursive} and Sentiment140 \cite{go2009twitter} datasets offer varying complexity levels. In the realm of natural language inference, the GLUE benchmark \cite{wang2018glue} tasks including MNLI \cite{williams2017broad}, QNLI \cite{rajpurkar2016squad}, and RTE \cite{dagan2005pascal} provide standardized evaluation frameworks. For cross-lingual scenarios, XGLUE-NC \cite{liang2020xglue}, XNLI \cite{conneau2018xnli}, and MasakhaNEWS \cite{adelani2023masakhanews} have proven particularly valuable for evaluating multilingual capabilities in linguistically diverse federated settings. Medical entity extraction represents a specialized classification domain with significant privacy implications, making it naturally suited for federated approaches. This task involves classifying tokens into predefined medical entity categories. The CBLUE \cite{zhang2022cblue} dataset provides standardized evaluation for Chinese biomedical language understanding. Recommendation systems in federated learning address the challenge of providing personalized suggestions while preserving user privacy. These systems often employ classification approaches to categorize items and predict user preferences based on textual descriptions. Datasets such as Amazon (Office-Arts) \cite{ni2019justifying} and OnlineRetail-Pantry \cite{chen2024pfcr} provide realistic recommendation scenarios with user-specific preferences.

Regarding foundation models for text classification, the BERT family has demonstrated strong baseline performance. BERT \cite{devlin2019bert}, DistilBERT \cite{sanh2019distilbert}, and RoBERTa \cite{liu2019roberta} serve as effective starting points for English text classification tasks, with DistilBERT being particularly suitable for resource-constrained clients. For multilingual scenarios, XLM-RoBERTa \cite{conneau2020unsupervised} and mBERT provide strong capabilities across languages. Larger models such as RoBERTa-Large and DeBERTa \cite{he2021deberta} offer enhanced performance but typically require more parameter-efficient approaches in federated settings due to communication constraints. 

Several federated PEFT methods have demonstrated promising results for text classification. Additive approaches such as FedAdapter and C2A have shown strong performance on news classification tasks with heterogeneous data distributions. Reparameterized methods including FFA-LoRA, RoLoRA, FedSA-LoRA, and PF2LoRA adapt Low-Rank Adaptation techniques to federated settings, proving particularly effective on GLUE benchmark tasks. For cross-lingual settings, specialized approaches like MFPT offer capabilities to handle language-based heterogeneity across clients. When addressing text classification in federated learning, the degree of data heterogeneity should be a primary consideration. For statistical heterogeneity where feature distributions vary across clients, methods like FRLoRA and SLoRA with RoBERTa models have demonstrated strong results. For label distribution skew, approaches such as GF-PT offer targeted solutions by adapting prompt-tuning techniques to handle imbalanced class distributions.

\subsubsection{Text Generation}
Text generation in federated settings has gained significant attention with the rise of large language models. This domain encompasses tasks ranging from summarization to open-ended generation, instruction following, and other specialized generation tasks, each presenting unique challenges in federated environments.

The dataset landscape for federated text generation is diverse and task-specific. For summarization tasks, QMSum \cite{zhong2021qmsum} provides domain-specific contexts (Academic, Committee, Product) that naturally fit federated scenarios where clients have data from different domains. For instruction-following capabilities, the Flan \cite{wei2022flan} dataset has emerged as a standard benchmark for evaluating federated LLM fine-tuning, particularly valuable for simulating heterogeneous task distributions across clients. Conversational datasets such as Dolly \cite{dolly2023}, Alpaca \cite{alpaca2023}, Wizard \cite{xu2023wizardlm}, and ShareGPT \cite{sharegpt2023} offer interaction data that can be partitioned by topic or domain for federated learning experiments. For evaluating complex reasoning capabilities, GSM-8K \cite{cobbe2021gsm8k} and Rosetta \cite{rosetta2023} provide mathematical and logical reasoning tasks. Machine translation in federated settings faces inherent challenges due to language diversity across clients. This text generation task involves producing accurate translations between language pairs while handling the inherent heterogeneity of multilingual data. The MNMT \cite{aharoni2019massively} datasets naturally fit federated learning scenarios, as they can be partitioned by language pairs to create realistic non-IID distributions.

Foundation models for text generation in federated settings span a range of scales and architectures. Smaller LLMs such as TinyLlama \cite{zhang2023tinyllama}, MobileLLaMA-1.4B \cite{mobilellama2023}, and PaLM 2 (XXS, XS) \cite{anil2023palm2} offer reasonable performance with lower computational requirements, making them suitable for resource-constrained federated scenarios. Medium-sized models including Llama-2 \cite{touvron2023llama2}, LLaMA \cite{touvron2023llama}, and Mistral \cite{jiang2023mistral} provide stronger performance while remaining manageable in federated settings with parameter-efficient approaches. For summarization-specific tasks, specialized architectures such as BART \cite{lewis2019bart} serve as effective baselines.

Several federated PEFT approaches have demonstrated promising results for text generation tasks. For scenarios with heterogeneous instruction distributions, methods such as FedDPA, FedALT, and FedOA using the Flan dataset have shown how to effectively handle diverse task types across clients. For domain-specific summarization, ADAFED has demonstrated strong results with additive tuning approaches. LoRA-based methods including FLoRA, HETLORA, and FlexLoRA have been successfully adapted to various foundation models, offering efficient solutions for federated LLM fine-tuning. The choice of approach for text generation in federated learning should consider both the computational capabilities of clients and the nature of data heterogeneity. For resource-constrained environments, approaches like FedP2EFT with smaller models provide efficient solutions, while scenarios with more capable clients can leverage methods like FlexLoRA with larger models such as LLaMA-3.

\subsubsection{Hybrid NLP Tasks}
Beyond the clear categories of text classification and text generation, some NLP tasks exhibit characteristics of both paradigms or serve specialized purposes in federated PEFT research.

Masked language modeling represents a versatile task with characteristics of both text classification and generation. It can serve both as a pretraining objective and as a means of adapting foundation models to domain-specific language. When predicting masked tokens, the model must both classify the most likely token from a vocabulary (classification) and generate contextually appropriate completions (generation). The WikiText-2 \cite{merity2016pointer} dataset provides a standardized benchmark, while Transformer-based architectures serve as foundations. Methods such as DepthFL employ selective tuning for efficient federated pretraining with masked language modeling objectives, demonstrating how this hybrid task can be effectively addressed in resource-constrained federated environments.

\begin{longtable}{>{\centering\arraybackslash}m{0.15\linewidth} >{\centering\arraybackslash}m{0.13\linewidth} >{\centering\arraybackslash}m{0.15\linewidth} >{\centering\arraybackslash}m{0.22\linewidth} >{\centering\arraybackslash}m{0.20\linewidth}}
    \caption{Federated Fine-tuning for NLP tasks}
    \vspace{10 pt}
    \label{tab: nlp} \\
    \hline
    \textbf{Applications} & \textbf{Method} & \textbf{Categories} & \textbf{Foundation Model} & \textbf{Experiment Dataset} \\
    \hline
    \endfirsthead
% 继续表头（翻页时重复）
    \hline
    \textbf{Applications} & \textbf{Method} & \textbf{Categories} & \textbf{Foundation Model} & \textbf{Experiment Dataset} \\
    \hline
    \endhead
    
    % 表格内容 - 按Application分组
    Text Classification & C2A & Additive & DistilBERT, XLM-RoBERTa & 20Newsgroup, XGLUE-NC \\
    \cline{2-5}
    & FedAdapter & Additive & BERT, DistilBERT & 20NEWS, AGNEWS, SEMEVAL \\
    \cline{2-5}
    & MFPT & Additive & XLM-RoBERTa & XGLUE-NC, XNLI, MasakhaNEWS \\
    \cline{2-5}
    & GF-PT & Additive & RoBERTa-Large & Sentiment140, CoLA, SST2, FDU-MTL \\
    \cline{2-5}
    & FedPETuning & Multiple categories & RoBERTa-Base & GLUE \\
    \cline{2-5}
    & FFA-LoRA & Reparameterized & RoBERTa-Large & MNLI, SST-2, QQP, QNLI \\
    \cline{2-5}
     & RoLoRA & Reparameterized & RoBERTa-Large, Llama-2-7B & SST-2, QNLI, MNLI, QQP, RTE \\
    \cline{2-5}
    & FedSA-LoRA & Reparameterized & RoBERTa-large & MNLI, SST2, QNLI, QQP, RTE \\
    \cline{2-5}
    & PF2LoRA & Reparameterized & BERT, RoBERTa, DeBERTa & MNLI, SST-2, QQP, QNLI, CoLA, SQuAD \\
    \cline{2-5}
    & FRLoRA & Reparameterized & RoBERTa-base, LLaMA-2-7B & RTE, COLA, QNLI, 20NEWS \\
    \cline{2-5}
    & SLoRA & Reparameterized & Albert, DistilBERT & 20News group, News Category \\
    \cline{2-5}
    & LEGEND & Reparameterized & RoBERTa, DeBERTa & SST-2, QNLI, QQP, MNLI \\
    \hline
    
    Text Generation & ADAFED & Additive & BART, LED, LongT5 & QMSum (Academic, Committee, Product) \\
    \cline{2-5}
    & FLoRA & Reparameterized & TinyLlama, Llama, Llama2 & Dolly, Alpaca, Wizard, ShareGPT \\
    \cline{2-5}
    & FedDPA & Reparameterized & LLaMA-7B, LLaMA-13B & Flan \\
    \cline{2-5}
    & FedALT & Reparameterized & LLaMA2-7B & Flan \\
    \cline{2-5}
    & FedOA & Reparameterized & LLaMA-7B & Flan \\
    \cline{2-5}
    & FedP2EFT & Reparameterized & mBERT, MobileLLaMA-1.4B, Llama-3.2-3B & XNLI, MasakhaNEWS \\
    \cline{2-5}
    & HETLORA & Reparameterized & PaLM 2 (XXS, XS) & Multi-Session Chat, Reddit Summarization \\
    \cline{2-5}
    & FlexLoRA & Reparameterized & DataJuicer (1.3B), LLaMA-3 (8B) & Natural Instructions, Dolly-15K \\
    \cline{2-5}
    & FedBiOT & AReparameterized & LLaMA-2-7B & GSM-8K, Rosetta, Dolly-15K \\
    \hline
    
    Machine Translation & Fed-MNMT & Additive & mBART-50 & MNMT \\
    \hline
    
    Medical Entity Extraction & MT-FedAdapter & Additive & BERT, BERT-wwm & CBLUE \\
    \hline
    
    Recommendation & PFCR & Additive & BERT & Amazon (Office-Arts), OnlineRetail-Pantry \\
    \hline
    
    Masked Language Modeling & DepthFL & Selective & Transformer & WikiText-2 \\
    \hline
    Arithmetic Reasoning & FedEx-LoRA & Reparameterized & Mistral-7B, Gemma-2 9B & GSM8K, MATH \\
    \cline{2-5}
    & FedSA-LoRA & Reparameterized & LLaMA3-8B & GSM8K \\
    \hline
\end{longtable}

\subsection{Computer Vision Tasks}
\subsubsection{Image Classification}
Image classification in federated settings encompasses a broad range of visual recognition tasks, from general object recognition to fine-grained categorization and cross-domain recognition. This domain presents particular challenges in federated learning due to potential variations in visual characteristics across clients.
The dataset landscape for federated image classification is rich and diverse. General classification benchmarks such as CIFAR-10/100 \cite{krizhevsky2009learning}, Caltech101 \cite{fei2004learning}, and SVHN \cite{netzer2011reading} provide standard evaluation frameworks with varying complexity levels. Fine-grained classification datasets including Flowers102 \cite{nilsback2008automated}, OxfordPets \cite{parkhi2012cats}, and Food101 \cite{bossard2014food} offer specialized categories that can simulate domain expertise across clients. 

Foundation models for image classification in federated environments have evolved significantly. CLIP \cite{radford2021learning}, with variants including ResNet \cite{he2016deep}, ViT \cite{dosovitskiy2021image}, offers strong visual-semantic understanding that transfers well across domains, making it particularly valuable for heterogeneous federated scenarios. Vision Transformers (ViT) provide powerful visual representations, while ResNet architectures still serve as lightweight baselines for resource-constrained scenarios. Alternative approaches such as MLP-Mixer \cite{tolstikhin2021mlp} offer competitive performance with different architectural paradigms.

Several federated PEFT methods have demonstrated promising results for image classification. Prompt-based approaches including pFedMoAP, Powder, FedAPT, pFedPrompt, and PromptFL adapt CLIP models to federated settings with minimal parameter updates, proving particularly effective for cross-domain generalization. Hybrid approaches such as FedRA combine selective and additive tuning for enhanced performance on challenging domains, while LoRA-based methods like LoRA-FAIR adapt reparameterization techniques to vision models in federated settings. The choice of approach for image classification in federated learning should consider the visual domain distribution across clients. 

\subsubsection{Advanced CV Tasks}
Beyond basic image classification, several advanced computer vision tasks have been addressed in federated PEFT research, presenting unique challenges and opportunities.

Domain adaptation explicitly addresses the challenge of transferring knowledge across visual domains, a common scenario in federated learning where clients have data from different sources. Datasets such as DomainNet \cite{peng2019moment}, PACS \cite{li2017deeper}, and Officehome \cite{gong2012geodesic} are designed specifically for evaluating domain adaptation capabilities, containing images from multiple distinct visual domains. Foundation models such as CLIP demonstrate strong zero-shot transfer capabilities across domains, while methods such as MPFT and DiPrompT adapt prompt-based tuning to domain adaptation in federated settings, explicitly modeling domain relationships for effective knowledge transfer.

Multimodal tasks in federated settings involve joint understanding of multiple modalities, such as vision and language. This domain encompasses visual question answering (using datasets such as GQA \cite{hudson2019gqa}, ScienceQA \cite{lu2022learn}, and OCRVQA \cite{mishra2019ocr}) and visual grounding (with datasets like COCO \cite{lin2014microsoft} and RefCOCO \cite{kazemzadeh2014referitgame}). Foundation models such as LLaVA \cite{liu2023visual}, which combines CLIP ViT-L/14 and Vicuna-7B \cite{chiang2023vicuna}, provide strong multimodal capabilities, while specialized architectures including ViLT \cite{kim2021vilt} and VAuLT \cite{chochlakis2022vault} offer different approaches to vision-language integration. Methods such as Pilot adapt multimodal models to federated settings with prompt-based tuning, addressing the complex heterogeneity inherent in multimodal tasks by allowing modality-specific adaptations.

\begin{longtable}{>{\centering\arraybackslash}m{0.15\linewidth} >{\centering\arraybackslash}m{0.13\linewidth} >{\centering\arraybackslash}m{0.15\linewidth} >{\centering\arraybackslash}m{0.22\linewidth} >{\centering\arraybackslash}m{0.20\linewidth}}
    \caption{Federated Fine-tuning for CV tasks}
    \vspace{10 pt}
    \label{tab:fl_peft_cv} \\
    \hline
    \textbf{Applications} & \textbf{Method} & \textbf{Categories} & \textbf{Foundation Model} & \textbf{Experiment Dataset} \\
    \hline
    \endfirsthead
    
    % 继续表头（翻页时重复）
    \hline
    \textbf{Applications} & \textbf{Method} & \textbf{Categories} & \textbf{Foundation Model} & \textbf{Experiment Dataset} \\
    \hline
    \endhead
    
    % 表格内容 - 按Application分组
    Image Classification & pFedMoAP & Additive & CLIP (ResNet50, ViT-b-16) & Flowers102, OxfordPets, Food101, DomainNet \\
    \cline{2-5}
    & Powder & Additive & ViT-B/16 & ImageNet-R, DomainNet, Office-Caltech10 \\
    \cline{2-5}
    & FedAPT & Additive & CLIP (ViT-B/32) & Office-Caltech10, DomainNet \\
    \cline{2-5}
    & pFedPrompt & Additive & CLIP (ViT-B/32) & Caltech101, Flowers102, OxfordPets, Food101, DTD, UCF101 \\
    \cline{2-5}
    & SGPT & Additive & ViT-B/16 & CIFAR-100, Five-dataset, Office-Caltech10, DomainNet \\
    \cline{2-5}
    & FedHPL & Additive & ViT-B/16, ViT-S/16, ViT-L/16 & CIFAR10, CIFAR100, SVHN \\
    \cline{2-5}
    & FedOTP & Additive & CLIP (ResNet50, ViT-B/16) & Food101, DTD, Caltech101, Flowers102, OxfordPets \\
    \cline{2-5}
    & PromptFolio & Additive & CLIP (ViT B/16, ResNet50) & CIFAR-10/100, DomainNet, Office-Caltech10 \\
    \cline{2-5}
    & PromptFL & Additive & CLIP & Caltech101, Flowers102, OxfordPets \\
    \cline{2-5}
    & pFedPG & Additive & ViT & Office-Caltech10, DomainNet, CIFAR-10/100, Dermoscopic-FL \\
    \cline{2-5}
    & FedRA & Selective and Additive & ViT, MLP-Mixer & DomainNet, NICO++ \\
    \cline{2-5}
    & FedPCL & Selective & ResNet18, ViT & Digit-5, Office-10, DomainNet \\
    \cline{2-5}
    & FedPEFT-Bias & Selective & ViT-Base & CIFAR-100, Resisc45, PCam, UCF101, HMDB51 \\
    \cline{2-5}
    & LoRA-FAIR & Reparameterized & ViT, MLP-Mixer & DomainNet, NICO++ \\
    \hline
    
    Domain Adaptation & MPFT & Additive & CLIP (ViT-B-32, ConvNeXT-Base) & DomainNet, PACS \\
    \cline{2-5}
    & DiPrompT & Additive & CLIP & PACS, Officehome, VLCS \\
    \hline
    
    Multimodal & Pilot & Additive & LLaVA 1.5 (CLIP ViT-L/14 + Vicuna-7B) & GQA, COCO, RefCOCO, ScienceQA, OCRVQA \\
    \cline{2-5}
    & FedDAT & Additive & ViLT, VAuLT & COCO, GQA \\
    \hline
\end{longtable}

\section{Future Directions} \label{future}

\subsection{Scalability to Larger Foundation Models}

As foundation models continue to scale exponentially—with recent architectures surpassing trillions of parameters \cite{achiam2023gpt}—the challenges of efficiently fine-tuning such models in federated settings become increasingly pronounced. While current PEFT methods demonstrate effectiveness for models with billions of parameters, their applicability and efficiency in the context of ultra-large models remain limited.

The inherent constraints of federated environments—where clients often have limited computational resources, storage capacity, and communication bandwidth—demand more sophisticated approaches for scaling PEFT to larger foundation models. Current federated PEFT methods face significant communication bottlenecks when dealing with extremely large models, as even the reduced parameter updates can become prohibitively large to transmit. Moreover, the memory footprint required to store and compute gradients for these models often exceeds the capabilities of many edge devices, effectively excluding them from the federated learning process.

Future research must address these federation-specific scaling challenges by developing techniques that consider the distributed nature of computation and heterogeneous client capabilities. Quantization-aware federated PEFT may represent promising direction, where both the foundation model weights and the adapter modules are quantized differently depending on client capabilities, with the server handling the necessary quantization conversions during aggregation. Communication-efficient aggregation algorithms specifically designed for federated PEFT with extremely large models could incorporate techniques such as adaptive precision, where different components of gradient updates are transmitted with varying precision based on their importance.

\subsection{Theoretical Analysis of PEFT in Federated Learning}

Despite promising empirical outcomes, the theoretical foundations of PEFT in federated learning are relatively underexplored compared to those of conventional FL methods. Strengthening these foundations is essential for principled algorithm design and robust deployment.

A primary direction is convergence analysis, which aims to establish theoretical guarantees for the convergence behavior of PEFT-based federated optimization algorithms, especially under client heterogeneity. This includes understanding how the reduced parameter update space in PEFT affects convergence rates compared to full-model fine-tuning. Similarly, deriving generalization bounds under non-IID data distributions would enhance confidence in PEFT’s applicability across diverse federated environments.

In addition, information-theoretic analysis can illuminate how well PEFT updates capture client-side information and how this influences cross-client generalization. Investigating the optimization landscape of PEFT in FL settings—especially the influence of low-rank or modular parameterizations on saddle points and local minima—may lead to better-informed optimization strategies. Formal frameworks linking parameter count, computational cost, and statistical efficiency are also needed to guide the selection of PEFT configurations under resource-constrained federated deployments.

\subsection{Sustainable and Green PEFT for Federated Learning}

With growing concern about the environmental impact of large-scale AI training \cite{bian2024cafe, li2023making}, there is an urgent need to develop sustainable and energy-efficient PEFT methods—particularly in federated settings where energy consumption is distributed across a multitude of devices.

Energy-aware PEFT methods that jointly optimize for parameter and energy efficiency represent a crucial step forward. Such approaches could incorporate dynamic adaptation of computational load based on device-specific energy availability, such as battery level or access to renewable energy. Standardized metrics for evaluating the carbon footprint of federated PEFT pipelines would further support sustainable development and responsible deployment.

Additionally, advancing efficient knowledge transfer mechanisms—where fine-tuned models or modules are reused across tasks—could reduce the frequency of retraining and thereby limit energy expenditure. Developing ultra-low-power PEFT techniques targeting IoT and edge devices would enable broader and greener participation in federated learning. Furthermore, intelligent scheduling algorithms that align training rounds with periods of low grid carbon intensity or surplus renewable energy could significantly reduce environmental impact.

Finally, moving toward a lifetime efficiency perspective—assessing the cumulative resource consumption of a PEFT method over its entire deployment cycle—can offer a more holistic view of sustainability. This includes initial fine-tuning, periodic adaptations, and long-term maintenance, offering a paradigm shift in how we evaluate the environmental cost of model customization strategies in federated learning.

\section{Conclusion} \label{conclusion}

This survey presented a comprehensive examination of Parameter-Efficient Fine-Tuning methods within Federated Learning, highlighting how these strategies enable scalable and privacy-preserving adaptation of foundation models in distributed environments. As large-scale models become increasingly prevalent across domains such as natural language processing and computer vision, the need for efficient fine-tuning mechanisms that account for the unique constraints of federated systems has never been more critical.

We categorized existing FL-PEFT methods into three primary groups—additive, selective, and reparameterized approaches—and systematically reviewed how each class addresses key challenges such as data heterogeneity, communication efficiency, computational constraints, and personalization. Moreover, we surveyed task-specific applications of PEFT within FL, providing insights into their deployment across diverse real-world scenarios. Despite the encouraging progress in this area, several fundamental challenges remain open. Current methods often face scalability bottlenecks when applied to ultra-large foundation models, lack theoretical convergence guarantees, and rarely consider the environmental impact of distributed training. This survey aims to serve as a foundational reference for researchers and practitioners interested in the intersection of federated learning and efficient model adaptation. As the field continues to evolve, we hope this work not only captures the current state-of-the-art but also catalyzes further innovation toward more effective, sustainable, and inclusive FL-PEFT systems.

\bibliographystyle{splncs04}
\bibliography{ref}

%%%%%%%%%%%%%

\end{document}